# Hyperspectral band selection using genetic algorithm and support vector machines for early identification of charcoal rot disease in soybean

Koushik Nagasubramanian[1#], Sarah Jones[2#], Soumik Sarkar[3], Asheesh K. Singh[2], Arti Singh[2*], Baskar Ganapathysubramanian[1,3,4*]

[1]Department of Electrical and Computer Engineering, Iowa State University, Ames, IA, USA

[2] Department of Agronomy, Iowa State University, Ames, IA, USA

[3] Department of Mechanical Engineering, Iowa State University, Ames, IA, USA

[4] Plant Sciences Institute, Iowa State University, Ames, IA, USA

**#-Equal Contributions** (These authors contributed equally)

**\* Correspondence** (Corresponding Authors)

**Abstract**

Charcoal rot is a fungal disease that thrives in warm dry conditions and affects the yield of soybeans and other important agronomic crops worldwide. There is a need for robust, automatic and consistent early detection and quantification of disease symptoms which are important in breeding programs for the development of improved cultivars and in crop production for the implementation of disease control measures for yield protection. Current methods of plant disease phenotyping are predominantly visual and hence are slow and prone to human error and variation. There has been increasing interest in hyperspectral imaging applications for early detection of disease symptoms. However, the high dimensionality of hyperspectral data makes it very important to have an efficient analysis pipeline in place for the identification of disease so that effective crop management decisions can be made. The focus of this work is to determine the minimal number of most effective hyperspectral bands that can distinguish between healthy and diseased specimens early on in the growing season for proper management of the disease. Healthy and diseased hyperspectral data cubes were captured at 3, 6, 9, 12, & 15 days after inoculation. We utilized inoculated and control specimens from 4 different genotypes. Each hyperspectral image was captured at 240 different wavelengths in the range of 383 – 1032 nm. We formulated the identification of best band combinations from 240 bands as an optimization problem. We used a combination of genetic algorithm as an optimizer and support vector machines as a classifier for identification of maximally effective band combinations. A binary classification between healthy and infected samples using six selected band combinations obtained a classification accuracy of 97% and a F1 score of 0.97 for the infected class. The results demonstrated that these carefully chosen bands are more informative than RGB images, and could be used in a multispectral camera for remote identification of charcoal rot infection in soybean.

## 1 Introduction

Soybean [*Glycine max* (L.) Merr.] is the major oilseed crop and the second major crop overall produced by the United States (USDA, 2016). Soybean is used to produce biofuel, cooking oil, soy foods, and animal feed, among many other uses, but the crop is threatened by over 100 diseases with 35 believed to be important pathogens affecting soybean yield (Boerema et al., 2016; Hartman and J. Rupe L. L. Domier, J. A. Davis, K. L. Steffey, 2015).

Charcoal rot is an economically critical disease that affects soybean, as well as 500 other plant species worldwide, and is caused by the fungal pathogen *Macrophomina phaseolina* (Tassi) Goid (Doupnik, 1993; Hartman et al., 2015; Koenning and Wrather, 2010). Infection is favored by warm (30-35 °C), dry, drought-like conditions but can cause up to 50% yield loss even in irrigated environments (Gupta et al., 2012; Mengistu et al., 2011; Meyer, 1974; Wyllie and Scott, 1988). Charcoal rot earned its common name from the gray-silver discoloration caused by microsclerotia formation in the vascular tissue and pith of lower stems and roots of infected plants (Gupta et al., 2012; Wrather and al., 2008). These microsclerotia are small dark survival structures that persist in the soil and plant debris after harvest and can act as an inoculum source for charcoal rot infection during the next growing season (Gupta et al., 2012; Hartman and J. Rupe L. L. Domier, J. A. Davis, K. L. Steffey, 2015; Romero Luna et al., 2017). Symptoms generally become visible at the R5 to R7 reproductive stages, or from early seed to early maturity, but can occasionally be seen earlier as reddish-brown lesions on the hypocotyl of seedlings. In more mature infected plants, a reddish-brown discoloration of the vascular tissue in the roots and lower stem generally precedes foliar symptom development when diseased plants may yellow, then wilt and prematurely senesce with dead leaves and petioles remaining attached to the stem(Gupta et al., 2012; Hartman and J. Rupe L. L. Domier, J. A. Davis, K. L. Steffey, 2015; Mengistu et al., 2007). Black microsclerotia on the above ground plant are first visible at the stem nodes and can be seen in the epidermal and sub epidermal tissue of plant stems as well as scattered on dry pods and seed of more mature plants (Gupta et al., 2012; Hartman and J. Rupe L. L. Domier, J. A. Davis, K. L. Steffey, 2015). Management of charcoal rot has proven to be difficult as no fungicides are available for control and more work needs to be done to research the potential of seed treatments (Hartman and J. Rupe L. L. Domier, J. A. Davis, K. L. Steffey, 2015; Romero Luna et al., 2017). In addition, crop rotation is not viewed as a viable strategy to manage infection, unless economically feasible crop rotation programs with non-host crops can be developed due to *M. phaseolina's* large host range which also includes other important agronomic crops such as corn, cotton, and sorghum (Short, 1980; Su et al., 2001) . Furthermore, no commercial soybean varieties are considered resistant, though a few varieties demonstrate moderate resistance (Mengistu et al., 2007, 2011, 2013; Paris, 2006; Pawlowski et al., 2015; Smith and Carvil, 1997). However, a GWA study across both field and greenhouse environments recently reported a total of 19 SNPs associated with charcoal rot resistance in soybean (Coser et al., 2017). While over 800 soybean lines have been evaluated for charcoal rot resistance, identification of resistant genotypes has been limited due to a *need for an accurate, rapid, and consistent method for disease assessment and classification* (Mengistu et al., 2007; Romero Luna et al., 2017).

**Current state of disease assessment and outlook:** Multiple methods, which are predominantly visual, have been proposed for assessing charcoal rot severity in the field and indoor environments. These methods include evaluation of the intensity or length of stem and root discoloration caused by microsclerotia formation, evaluation of the percent chlorosis and necrosis of the plant canopy throughout the growing season, chlorosis and necrosis of foliage that remains attached to the plant at R7, calculation of colony forming unit index to quantify the microsclerotia content in the stem and root, and lesion length measurements of cut-stem inoculations on young plants (Barratt and Horsfall, 1945; Mengistu et al., 2007; Smith and Carvil, 1997; Twizeyimana et al., 2012). However, visual rating methods can be subjective and are susceptible to human error caused by rater ability, and inter/intra-rater reliability (Bock et al., 2010; Bock and Nutter Jr., 2011; James, 1974; Nutter, 1993).

Furthermore, visual ratings only take advantage of visible wavelengths of the electromagnetic spectrum. Hyperspectral imaging can capture both spectral and spatial information from a wider range of the electromagnetic spectrum including the visible and near-infrared regions. Automating disease severity rating through hyperspectral imaging offers a potential solution to the standardization and reliability issues in current visual rating systems. Extraction of pure reflectance spectra from each pixel at different wavelengths enables one to relate changes in reflectance values to disease symptoms (Mahlein et al., 2012a, 2012b). Recent plant pathology and phenotyping studies have utilized hyperspectral imaging data to study the effect of different pathogens. Examples include approaches to identify differences in the reflectance patterns of resistant and susceptible barley genotypes inoculated with powdery mildew; the content of charcoal rot (*M. phaseolina*) microsclerotia in ground root and stem tissue as a method for rating infection severity, and hyperspectral imaging to distinguish between the symptoms of Cercospora leaf spot, powdery mildew, and leaf rust at different developmental stages in sugar beet (Fletcher et al., 2014; Kuska et al., 2015; Mahlein et al., 2012b; Mirwaes et al., 2016).

A key issue with utilizing hyperspectral imaging is that the resulting hyperspectral data cubes are high dimensional and contain redundant information which reduces the ability to distinguish between classes. Using a hyperspectral camera on a drone for crop disease identification and phenotyping can also generate large quantities of data during the flight making it necessary to have a large on-board storage capacity and also substantially increases computational cost for any subsequent analysis. Therefore, there is a need to develop an analysis pipeline to reduce dimensionality of the data and to select optimal wavelengths that are most useful for phenotyping and disease identification. This serves as the motivation of this study.

Feature extraction and feature selection are two different methods for dimensionality reduction of hyperspectral data. Feature extraction methods such as Principal Component Analysis (PCA), Linear Discriminant Analysis (LDA), Independent Component Analysis (ICA) and Maximum Noise Fraction (MNF) project the original hyperspectral feature space into a new low-dimensional feature space (Bandos et al., 2009; Liu et al., 2009; Tyo et al., 2003; Villegas-Fernández et al., 2011). Feature extraction methods alter the physical meaning of the hyperspectral data during transformation to a new (and lower) dimensional space whereas feature selection methods preserve the original features. Feature selection essentially boils down to carefully selecting a subset of the available bands (i.e. band selection) that preserves certain traits of the full dataset. Feature selection methods are broadly classified into supervised or unsupervised methods. Some supervised band selection methods use class separability metrics like Euclidean distance, transformed divergence, Bhattacharyya distance, Jeffreys–Matusita (JM) distance (Keshava, 2002; Yang et al., 2011). A band selection method based on estimation of mutual information for classification of hyperspectral images was studied by Guo et al., 2006. Sequential search strategies like Sequential Forward Selection (SFS), Sequential Floating Forward Selection (SFSS), Sequential Backward Selection (SBS) and Sequential Backward Floating selection (SBSS) have also been used for band selection (Pudil et al., 1994; Serpico and Bruzzone, 2001). These sequential search algorithms are simple and suboptimal. Evolutionary methods such as Particle Swarm Optimization (PSO) and Genetic Algorithms (GA) which can search for global optimal solutions have been found to be successful in effective band selection (Li et al., 2011; Yang et al., 2007). In this study, we use an evolutionary method, specifically GA, as an optimizer along with Support Vector Machine (SVM) as a classifier for effective band selection. GA-SVM based model have been successful in band selection for classification of remotely sensed hyperspectral images (Bazi and Melgani, 2006; Chen et al., 2014; Li et al., 2011; Pal, 2006; Zhuo et al., 2008). Although computationally costly, evolutionary algorithms can give better optimal solution than sequential algorithms since the best feature combination is selected simultaneously.

The objectives of this study were hyperspectral imaging enabled early identification of charcoal rot disease, and to determine the most effective minimum number of bands for discrimination of healthy and charcoal rot infected stems. This study shows that a Genetic Algorithm-Support Vector Machine based model can be used in selecting the most effective band combination for early detection of charcoal rot disease in soybeans. Additionally, using F1-Score as an optimization metric instead of classification accuracy can overcome the skewness of classification accuracy metric for the dominant class of an imbalanced dataset (number of healthy samples more than the number of infected samples). An effective six band combination for discrimination of healthy and charcoal rot infected stems was found. Early identification of charcoal rot disease at three days after inoculation was possible using the selected waveband combination.

## 2      Materials and Method

### 2.1    Plant Material

Four soybean genotypes, Pharaoh (susceptible), PI479719 (susceptible), DT97-4290 (moderately resistant), and PI189958 (moderately resistant) were included in this study. Two seed of each genotype were planted in a commercial soil substrate (Sungro horticulture professional growing mix) in 8 oz styrofoam cups in a growth chamber at 30°C day/21°C night with a 16-hour photoperiod. Each styrofoam cup was supplemented with 1/8tsp (0.65g) of osmocote 15-9-12 at planting. Ten days after planting, plants were thinned down to one plant per pot choosing the most vigorous plant. Plants were arranged in a randomized complete block design with four replications. The two treatments were inoculation and mock-inoculation, and data collection was done at three day intervals [3, 6, 9, 12, and 15 days after inoculation (dai)]. Replication 1 was planted in the growth chamber in September 2016 and imaged at 3, 6, 12, & 15 dai. Replications 2-4 were planted in November 2016. Lesion length ratings and data cubes were collected at 3, 6 and 9 dai in replications 2-4 focusing on early disease development time points.

### 2.2    Culture and Inoculation of *M.phaseolina*

The pathogen *M. phaseolina* 2013X, originally collected from the field in Iowa in 2013, was re-isolated from inoculated stems of soybean plants grown in the growth chamber. Four days before inoculation (17 days after planting), 0.5cm plugs of *M. phaseolina* were transferred to Potato Dextrose Agar (PDA) plates which were then stored in the dark at 30°C for four days. Twenty-one days after planting, the four soybean genotypes were inoculated according to the cut-stem inoculation technique (Twizeyimana et al., 2012). Sterile 200ul pipette tips were placed open end down into the media around the leading edge of the fungal colony. Plant stems were severed 40mm above the unifoliate node and a pipette tip carrying a disk of uncontaminated PDA media for the mock-inoculation treatment and PDA media + *M. phaseolina* mycelia for the inoculation treatment, was placed onto the cut stem and the open wound imbedded in the media. Three days after inoculation, pipette tips were removed from all plants.

## 2.3 Hyperspectral Image Acquisition

Pika XC hyperspectral line scanning imager (Resonon, Bozeman, MT) was used for imaging, and hyperspectral data cubes were constructed. The Pica XC camera has a spectral resolution of 2.5 nm, with 240 spectral channels covering a spectral range from 382-1032 nm. Hyperspectral images of healthy and charcoal rot infected stems were collected at different time points, as explained previously, for classification.

The imaging system also includes a mounting tower, linear translation stage, and a computer pre-loaded with SpectrononPro software (Resonon, Bozeman, MT). Illumination was provided by two 70-watt quartz-tungsten-halogen Illuminator lamps (ASD Inc., Boulder, CO) which provide stable illumination over a 350 – 2500 nm range. The distance between the lamps and the plant stem being imaged was 54 cm with lights pointed towards the sample at a 45-degree angle. Prior to imaging, the ASD pro-lamps were turned on and warmed up for at least 20 minutes to produce a stable light source.

Using the SpectrononPro software interface, the camera exposure was set automatically and focus adjusted manually using a lens of f-number (ratio of focal length and diameter of a lens) of $f/1.4$. The system was then calibrated to a white reference tile and a dark reference with the lens cap covering the objective lens. Aspect ratio was adjusted using a concentric circles sheet provided by Resonon. Data was captured with reflectance values between 0 and 1. Figure 1 shows the hyperspectral imaging setup used in the study.

Plant stems were destructively imaged at each time point (3, 6, 9, 12 and 15 dai). All leaves were removed from the plant stem and the stem cut off at the soil surface just before hyperspectral data cube collection. Stems were placed on the linear translation stage for imaging with the longest edge of the lesion facing the camera lens and data cubes collected after system calibration. The hyperspectral data cube and corresponding RGB image were saved on an external hard drive.

**Figure 1**. Imaging system comprised of Resonon Pika XC hyperspectral camera and ASD Inc. quartz-tungsten-halogen Illuminator lamps.

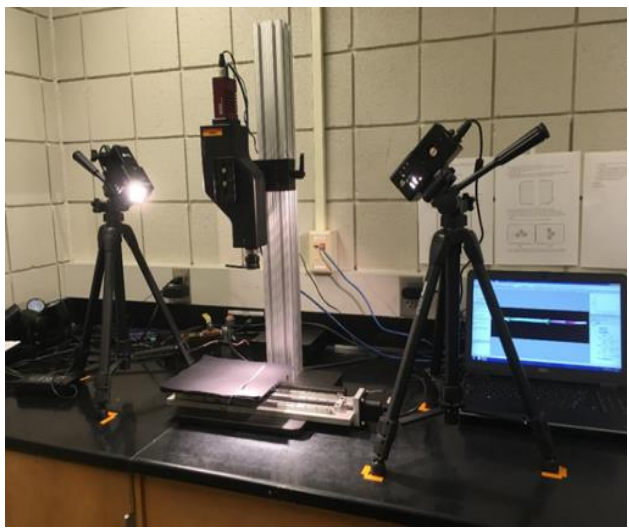

## 2.4 Charcoal rot rating protocol

In addition to stem images, disease progression was manually rated by measuring length (mm) of the exterior lesion, interior lesion, and dead tissue lesion. The exterior lesion was clearly visible as a reddish-brown to black discoloration proceeding from the inoculated end of the stem. The interior lesion, a reddish-brown discoloration of the vascular tissue, progressed farther than the exterior reddish-brown lesion and was measured to the lowest point of the dark reddish continuous discoloration from the inoculated end of the stem. Tissue death was the last symptom to develop and as such, the dead tissue lesion was shorter than the interior and exterior lesions and was measured to the extent of the dry, dead plant tissue. Measurement protocol was designed based on Twizeyimana et al., 2012, where charcoal rot lesion length was measured from the unifoliate node to the lowest edge of the lesion being measured. Figure 2 shows the interior and exterior and dead tissue lesion lengths of an infected soybean stem.

**Figure 2**. Charcoal rot disease ratings were obtained by measuring three different lesion elements of symptom development including the exterior lesion, dead tissue, and interior lesion length (mm).

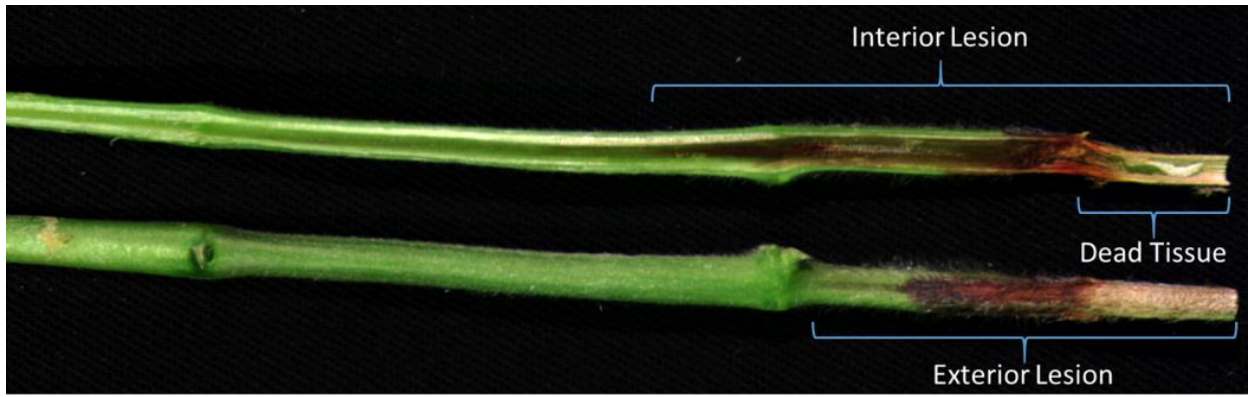

**Genetic Algorithm-Support Vector machine based feature selection**

## 2.5 Problem Definition

The identification of best band combination for maximally discriminating healthy and charcoal rot infected stems from a set of 240 bands was formulated as an optimization problem. A genetic algorithm (GA) based optimization protocol using support vector machine (SVM) as a classifier was used to find the most optimal bands for designing a multispectral camera system for phenotyping and disease identification. Spectral and spatial information from the hyperspectral images were used for early identification and classification of disease. The objective of the optimization was to find the best band combination that maximizes the classification performance (i.e. find the best k band combination that produces the best classification performance when distinguishing between healthy and diseased specimens). Figure 3 shows the flowchart of the GA-SVM architecture for band selection. MATLAB R2017a was used to implement the GA-SVM model.

**Figure 3**. GA-SVM architecture for selection of optimal bands

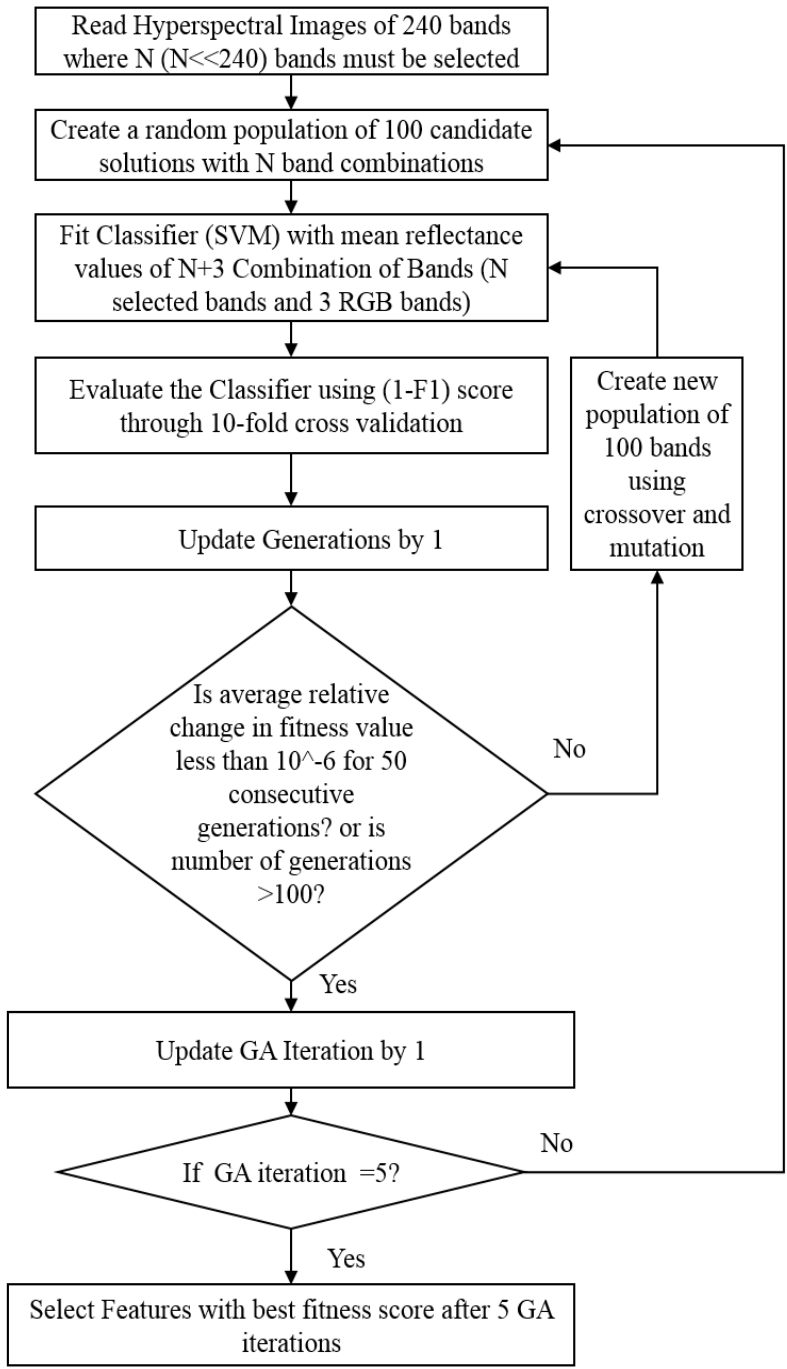

## 2.6 Support Vector Machine

Support Vector Machine is a kernel-based discriminative supervised learning algorithm for classification (Burges, 1998; Cortes and Vapnik, 1995). SVM is one approach for constructing a classifier that maps an input data (of N band information) to a class (healthy vs infected). SVM has been used with significant success in a variety of plant science applications (Naik et al., 2017; Singh et al., 2016). Formally, an SVM projects the data which are not separable linearly into a higher dimensional space using a kernel. It separates the classes with an optimal hyperplane that maximizes the margin between the classes (Boser, et al., 1992). In this study, we used Radial Basis Function (RBF) kernel to learn the non-linear classifier. SVM has been used as a classifier in wrapper based feature selection methods for classification of hyperspectral images (Bazi and Melgani, 2006; Bor-Chen et al., 2014; Covert et al., 2007; Li et al., 2011; Pal, 2006; Samadzadegan et al., 2012; Vaiphasa et al., 2007; Zhuo et al., 2008). After trial and error, the two Radial Basis Function (RBF) kernel parameters C and $\gamma$ were set to 1000 and 1, respectively.

## 2.7 Genetic Algorithm

Genetic Algorithms are population based stochastic search optimization techniques inspired by natural selection and natural genetics principles (Goldberg and Holland, 1988). The population of candidate solutions (i.e. bands) is represented as a long string of bits and is called 'chromosome'. Each of these chromosomes is assigned a score using a fitness function for evaluation (Goldberg, 1989). In this particular case, the fitness function evaluates how well the chromosome (i.e. that particular selection of bands) performs to distinguish between diseased and healthy specimens. These chromosomes are evolved in successive generations using selection, mutation and crossover genetic operators for exploring the solution space, until a best solution is obtained or termination criteria is encountered. Selection of chromosomes for reproduction can be done in diverse ways (Goldberg and Deb, 1991). One of the ways is to choose the pair of chromosomes in the population that provides relatively good fitness scores to perform crossover. Crossover operator randomly combines genetic information of two chromosomes. Mutation operator modifies some component of a chromosome to form random new populations in the search space which prevents GA from choosing local optimal solutions. The "elite" parameter decides the number of most-fit individuals passed from one generation to the next generation without changing. This process of selection, mutation and crossover is repeated for multiple generations to improve the population fitness.

It is important to carefully choose a well-defined and appropriate fitness function. After exhaustive numerical tests and exploration, we chose the F1 score of the infected class as a useful tool to evaluate performance of the classifier. Maximizing only precision or recall does not imply good classification performance. F1 score is defined as harmonic mean of precision and recall values providing equal weightages to both precision and recall scores. A good F1 score is also indicative of good classification performance. Equations 1, 2, and 3 provide the formulas for precision, recall, and F1 score metrics. The value of F1 score can vary from 0 to 1. A value of 1 and 0 is obtained for best and worst classification performance respectively. The fitness value was chosen as (F1 score) of the 10-fold cross validated SVM classification result. F1 score is a better metric over classification accuracy for measuring the classification performance of an imbalanced data, as classification accuracy is a biased metric which favors the class with more samples (healthy samples in our case). The objective of the GA was to find the best band combination that maximizes the F1 score. Table 1 shows the variables of the confusion matrix to analyze the performance of the classification.

**Table 1. Confusion Matrix definition**

|  | Predicted Class | |
|---|---|---|
| True Class ↓ | **Infected** | **Healthy** |
| **Infected** | True positive (TP) | False Negative (FN) |
| **Healthy** | False Positive (FP) | True Negative (TN) |

$$Precision = \frac{TP}{TP + FP} \quad (1)$$

$$Recall = \frac{TP}{TP + FN} \quad (2)$$

$$F1\ Score = \frac{2 * Precision * Recall}{(Precision + Recall)} \quad (3)$$

The termination criteria depend on the average change in fitness value for 50 continuous generations or the maximum number of generations allowed which were 100 in our study. The last generation of GA iteration will contain the most optimal solution.

We choose to augment the hyperspectral bands with some visible spectrum (RGB information). We do this since RGB cameras are inexpensive, light weight, and can be attached to drones easily for capturing images. Therefore, the input feature to the SVM classifier consists of a fixed part and variable part. The mean values of reflectance from three wavelengths 475.56 nm, 548.91 nm and 652.14 nm representing red, green and blue colors respectively were used as fixed part of the input feature. The variable part of the input feature was chosen by the GA. The input chromosome comprises of bits each representing one of the total 240 bands of the input hyperspectral image. The number of bits in a chromosome is equal to the total number of bands to be selected by the GA. The number of bits chosen were 3 in our study. In total, the input features consisted of six wavelengths, including RGB and the wavelengths selected by the GA. Binary tournament,

Laplace, and power methods were used for selection, crossover and mutation respectively. Table 2 provides the implementation details of the GA.

**Table 2. Implementation details of Genetic Algorithm.**

| Parameters | |
|---|---|
| Number of Genetic Algorithm iterations | 5 |
| Population | 100 |
| Maximum Number of generations | 100 |
| Crossover Probability | 0.8 |
| Elite Count | 2 |
| Mutation Probability | 0.2 |
| Selection | Binary selection tournament |
| Crossover | Laplace crossover |
| Mutation | Power mutation |
| Stopping Criteria | Average change in best fitness value is less than $10^{-6}$ for 50 generations or number of generations=100 |

## 2.8 Data pre-processing

The dataset contains 111 hyperspectral images of size 500 x 1600 pixels. Replications 1-4 provided 39, 24, 24, and 24 data cubes respectively. Data cubes from each replication were distributed among the training and testing datasets. Seventy-two hyperspectral images were used for training and 39 hyperspectral images were used for testing. The training set had 35 data cubes of healthy stems and 37 data cubes of diseased stems. The testing set had 21 data cubes of healthy stems and 18 data cubes of diseased stems. Since the number of test data was small, to increase the amount of data for developing the model and for prediction of disease progression to get a better understanding of severity of the disease spread, each of the hyperspectral stem images was divided into patches of size 500 x 64 pixels for training and testing purpose. The healthy (mock-inoculated) and diseased (inoculated) samples allowed for testing and training for classification of diseased compared to healthy tissue. Training data was labeled using ground truth data of the measured interior lesion length (mm). A summary of the ground truth data for interior lesion length as well as the exterior and dead tissue lesion lengths can be seen in Table 3. Due to destructive nature of data collection, the expected trend of lesion length increasing over time is not always observed. The interior lesion length, measured in mm on the interior of the stem, was used for ground truth labelling of the image patches. A stem is determined as infected if at least one of the image patches of the stem is predicted as infected.

**Table 3. Mean and standard error of the mean for lesion length ratings (exterior, interior, and dead tissue) from the 3 earliest time points of lesion rating.**

| Trait | Time point | Genotype | Number of samples | Mean (mm) | Std Err Mean |
|---|---|---|---|---|---|
| Exterior | 3 DAI | DT97-4290 | 4 | 31.5 | 8.5 |
| | | Pharoah | 4 | 28.0 | 4.7 |

|  |  |  | | |
|---|---|---|---|---|
| | | PI189958 | 4 | 25.5 | 4.5 |
| | | PI479719 | 4 | 18.0 | 3.7 |
| | **6 DAI** | DT97-4290 | 4 | 31.0 | 7.1 |
| | | Pharoah | 4 | 28.5 | 4.4 |
| | | PI189958 | 4 | 28.5 | 2.5 |
| | | PI479719 | 4 | 22.8 | 2.3 |
| | **9 DAI** | DT97-4290 | 3 | 34.3 | 6.2 |
| | | Pharoah | 3 | 39.7 | 5.8 |
| | | PI189958 | 2 | 20.0 | 1.0 |
| | | PI479719 | 3 | 36.0 | 4.0 |
| **Interior Lesion Length** | **3 DAI** | DT97-4290 | 4 | 29.0 | 7.0 |
| | | Pharoah | 4 | 35.0 | 2.1 |
| | | PI189958 | 4 | 30.0 | 3.0 |
| | | PI479719 | 3 | 46.0 | 9.6 |
| | **6 DAI** | DT97-4290 | 4 | 37.5 | 6.3 |
| | | Pharoah | 4 | 49.8 | 9.5 |
| | | PI189958 | 4 | 34.3 | 3.6 |
| | | PI479719 | 4 | 26.5 | 6.8 |
| | **9 DAI** | DT97-4290 | 3 | 68.3 | 12.3 |
| | | Pharoah | 3 | 61.0 | 10.7 |
| | | PI189958 | 3 | 41.0 | 2.5 |
| | | PI479719 | 3 | 66.3 | 12.4 |
| **Dead Lesion Length** | **3 DAI** | DT97-4290 | 4 | 17.3 | 6.6 |
| | | Pharoah | 4 | 20.3 | 5.5 |
| | | PI189958 | 4 | 18.3 | 2.5 |
| | | PI479719 | 3 | 23.3 | 0.9 |
| | **6 DAI** | DT97-4290 | 4 | 25.0 | 6.4 |
| | | Pharoah | 4 | 22.8 | 5.0 |
| | | PI189958 | 4 | 16.0 | 1.8 |
| | | PI479719 | 4 | 16.8 | 3.0 |
| | **9 DAI** | DT97-4290 | 3 | 32.3 | 5.7 |
| | | Pharoah | 3 | 32.3 | 4.9 |
| | | PI189958 | 3 | 12.0 | 4.6 |
| | | PI479719 | 3 | 28.7 | 5.2 |

## 3     Results and Discussion

### 3.1    Spectral Reflectance

Figure 4 shows an example of mean reflectance curves of healthy and infected samples at various stages. It is seen that the maximum reflectance value of infected samples is less than the healthy sample and the trends of all infected samples looks similar. The reflectance value decreases as the severity of the charcoal rot disease increases.

**Figure 4.** Mean spectral reflectance curves of healthy and infected stems.

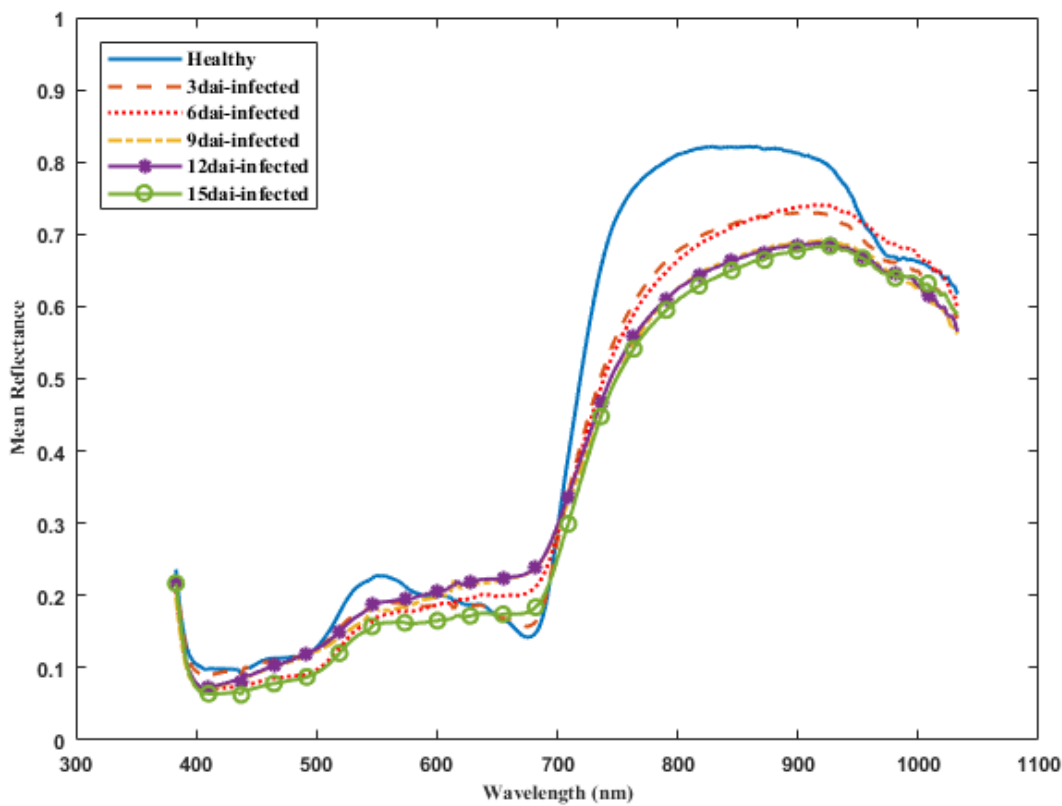

## 3.2 Feature Selection

The number of bands used for classification were reduced from 240 to 6 using our GA-SVM model. Table 4 shows the maximally effective 6 band combinations selected by the GA-SVM model including RGB bands.

**Table 4. Selected best band combination by Genetic Algorithm**

| Band combination | RGB wavelengths and three wavelengths selected by Genetic Algorithm |
| --- | --- |
| 6 | 475.56(B), 548.91(G), 652.14(R), 516.31, 720.05, 915.64 |

Binary classification results for selected wavelength combinations are shown in Table 5. The F1 score of the infected class and overall classification accuracy were 0.769 and 76.92% respectively using only RGB wavelengths whereas classification accuracy of 97% and F1-Score of 0.97 for 39 test stems were obtained using the selected 6 band combinations of GA.

**Table 5. Classification results**

| Band Combination | Confusion Matrix | | Precision | Recall | F1-Score | Healthy** | Infected** | Overall Accuracy (%) |
|---|---|---|---|---|---|---|---|---|
| 3(RGB) | TP=17 | FP=8 | 0.68 | 0.94 | 0.79 | 92.85 | 68 | 76.92 |
|  | FN=1 | TN=13 |  |  |  |  |  |  |
| 6 | TP=18 | FP=1 | 0.94 | 1 | 0.97 | 100 | 94 | 97 |
|  | FN = 0 | TN=20 |  |  |  |  |  |  |

**Per class accuracy (%)

The RGB wavelengths alone did not perform well, which might be because of their inability to differentiate between the reflectance values of a healthy stem and charcoal rot infected stem. The classification accuracy and F1 score of the selected 6 waveband combinations indicate that they were good at distinguishing between healthy and charcoal rot infected samples.

### 3.3 Early disease detection for 3-dai samples

The ability to detect disease early is very important for mitigation. Among 39 test stems, 11 were collected at 3-dai. Out of 11, 6 represent healthy stems and 5 were infected. The binary classification results for 3-dai samples are shown in Table 6. The classification accuracy and F1-score were 81.82% and 0.83 respectively using RGB wavelengths whereas the classification accuracy and F1 score were 90.91 and 0.90 respectively using the 6 band combinations. These results indicate that the specific wavelengths chosen in the six band combinations are responsive to disease symptoms even at the early stage of infections.

**Table 6. Classification results for 3-dai samples.**

| Band Combination | Confusion Matrix | | Precision | Recall | F1 | Healthy** | Infected** | Overall Accuracy (%) |
|---|---|---|---|---|---|---|---|---|
| 3(RGB) | TP=5 | FP=2 | 0.71 | 1 | 0.83 | 100 | 71.43 | 81.82 |
|   |            | FN=0   | TN=4 |   |      |     |       |       |
| 6 | TP=5       | FP=1   | 0.83 | 1 | 0.90 | 100 | 83.33 | 90.91 |
|   | FN=0       | TN=5   |      |   |      |     |       |       |

**Per class accuracy (%)

### 3.4 Disease length prediction

Identification of charcoal disease length progression is important for understanding the severity of the disease and helpful in understanding the resistance of various soybean genotypes to the disease. Figure 5 shows the predictions for each patch in an inoculated stem.

**Figure 5.** Prediction of stem patches by selected optimal wavelengths

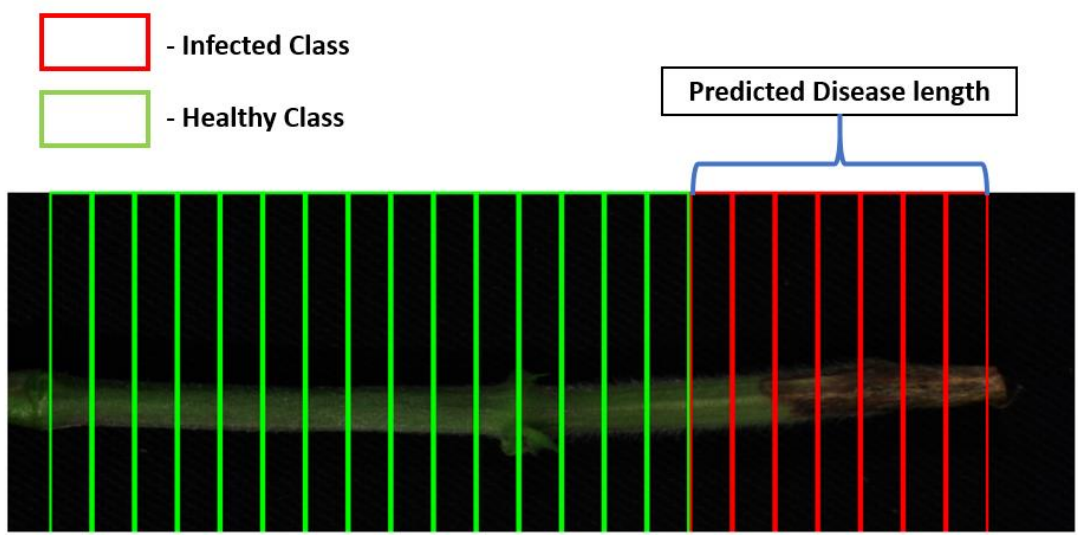

The total disease length is the distance from the inoculation point to the end of the farthest patch which was predicted as infected from the inoculation point. The total predicted disease length could be calculated by summing the length of the number of patches in one stem data cube classified as diseased. The predicted disease lengths for 39 test stems are shown in Figure 6. The disease length prediction for stem number 30 was incorrect due to misclassification of a patch at the end of the stem. For other stem samples, the predicted disease length was proportional to the interior lesion length.

**Figure 6.** Actual disease progression length (mm) compared to predicted disease progression length based on patch wise classification results.

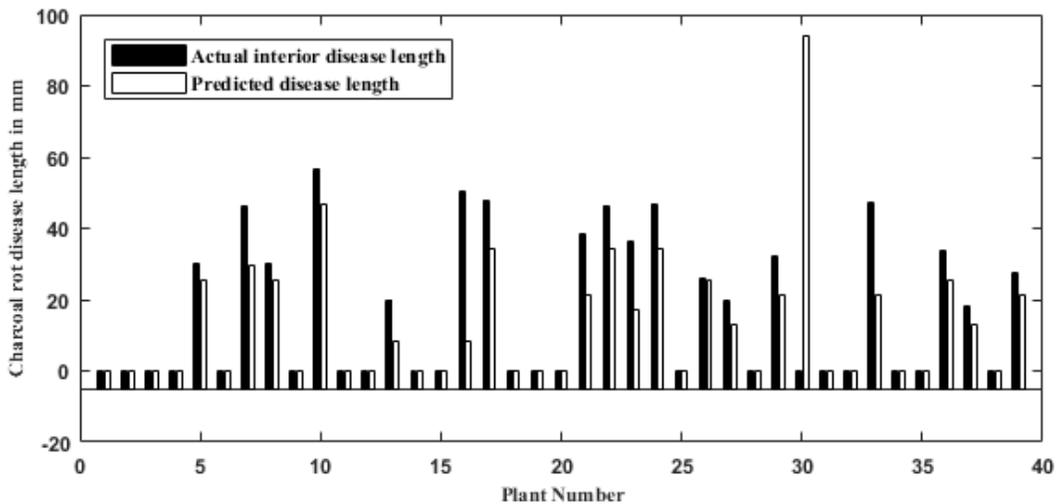

## 4   Conclusions

Hyperspectral images of four different soybean genotypes (two susceptible and two moderately resistant), half healthy and half infected with charcoal rot disease were collected at 5 different time points post infection. The main objectives of this study were to identify the most effective minimal number of bands from a set of 240 hyperspectral bands that are required for identification of charcoal rot disease and to analyze the performance of these bands in early detection of the disease.

The study used both spectral and spatial information (mean value of reflectance from different wavelengths) for disease identification. Due to imbalanced dataset of healthy and infected stems used in our study, the SVM classification performance which was optimized using GA for optimal wave band selection was evaluated for maximizing the F1 score value of the infected class instead of overall classification accuracy.

An effective six band combination for discrimination of healthy and charcoal rot infected stems was found. Early identification of charcoal rot disease at three days after inoculation was possible using the selected waveband combinations. The GA-SVM model obtained F1-score of 0.97 and classification accuracy of 97% using selected 6 hyperspectral band combinations. These results were better than those obtained using only the visible RGB wavelengths highlighting the importance of including the additional non-visible wavelengths for disease detection. The F1-score and classification accuracy for early detection (3-dai samples) samples were 0.90 and 90.91% respectively using the selected 6 bands. Two out of the three wavelengths selected (720.05nm, 915.64nm) along with the RGB bands in the six band combinations were selected in the near-infrared region and one was selected in the visible region (516.31nm) indicating that both near infrared region and visible region were useful in early identification of charcoal rot disease.

Genotypes with susceptible and moderately resistant responses to charcoal rot were used in this study. The length of disease progression (mm) in each stem was measured to understand the severity of the disease spread among different genotypes. Using hyperspectral imaging combined with GA-SVM enabled band selection resulting in a higher classification accuracy compared to visible wavelengths alone. However, this study focused on indoor imaging so future work should utilize field inoculations and evaluations to expand this technology into the field. Furthermore, field inoculations of diverse soybean genotypes will be imaged using a multispectral camera with the selected wavebands from the GA-SVM model for early identification of charcoal rot disease to understand the disease resistance of specific genotypes. Also, the length of disease progression in different genotypes will be studied with larger sample size to characterize their disease resistance. In all, this study provides an efficient methodology for selecting the most valuable wavebands from hyperspectral data to be used for early disease detection.

**Acknowledgements**

This work was funded by Iowa Soybean Association (AS), NSF/USDA NIFA grant (SS, BG, AS, AKS), ISU Research grant through the PIIR award (AS, AKS, SS, BG), Monsanto Chair in Soybean Breeding at Iowa State University (AKS) and PSI Faculty Fellow award (BG)